\pdfoutput=1

\documentclass[11pt]{article}

\usepackage[preprint]{acl}

\usepackage{times}
\usepackage{latexsym}

\usepackage[T1]{fontenc}

\usepackage[utf8]{inputenc}

\usepackage{microtype}

\usepackage{inconsolata}

\usepackage{graphicx}
\usepackage{amsmath}
\usepackage{amssymb}
\usepackage{mathtools}
\usepackage{amsthm}
\usepackage{bbm}
\usepackage{algorithm}
\usepackage{algorithmic}
\usepackage{enumitem}
\usepackage{multirow}
\usepackage{booktabs}
\usepackage[table, dvipsnames]{xcolor}
\usepackage[most]{tcolorbox}
\usepackage{fontawesome}
\usepackage[normalem]{ulem}
\useunder{\uline}{\ul}{}
\usepackage{array}
\usepackage{pifont}
\newcolumntype{C}[1]{>{\centering\arraybackslash}p{#1}}

\title{Chaining the Evidence: Robust Reinforcement Learning for Deep Search Agents with Citation-Aware Rubric Rewards}

\author{
Jiajie Zhang$^{1}$\thanks{Work was done when JZ interned at Zhipu AI.},
Xin Lv$^{2}$,
Ling Feng$^{1}$,
Lei Hou$^{1}$,
Juanzi Li$^{1}$ \\
$^{1}$Tsinghua University, $^{2}$Zhipu AI
}

\begin{document}

\maketitle

\begin{abstract}
Reinforcement learning (RL) has emerged as a critical technique for enhancing LLM-based deep search agents. However, existing approaches primarily rely on binary outcome rewards, which fail to capture the comprehensiveness and factuality of agents' reasoning process, and often lead to undesirable behaviors such as shortcut exploitation and hallucinations. To address these limitations, we propose \textbf{Citation-aware Rubric Rewards (CaRR)}, a fine-grained reward framework for deep search agents that emphasizes reasoning comprehensiveness, factual grounding, and evidence connectivity. CaRR decomposes complex questions into verifiable single-hop rubrics and requires agents to satisfy these rubrics by explicitly identifying hidden entities, supporting them with correct citations, and constructing complete evidence chains that link to the predicted answer.
We further introduce \textbf{Citation-aware Group Relative Policy Optimization (C-GRPO)}, which combines CaRR and outcome rewards for training robust deep search agents. 
Experiments show that C-GRPO consistently outperforms standard outcome-based RL baselines across multiple deep search benchmarks. Our analysis also validates that C-GRPO effectively discourages shortcut exploitation, promotes comprehensive, evidence-grounded reasoning, and exhibits strong generalization to open-ended deep research tasks. Our code and data are available at \url{https://github.com/THUDM/CaRR}.

\end{abstract}

\begin{figure}[!t]
    \centering
    \includegraphics[width=\linewidth]{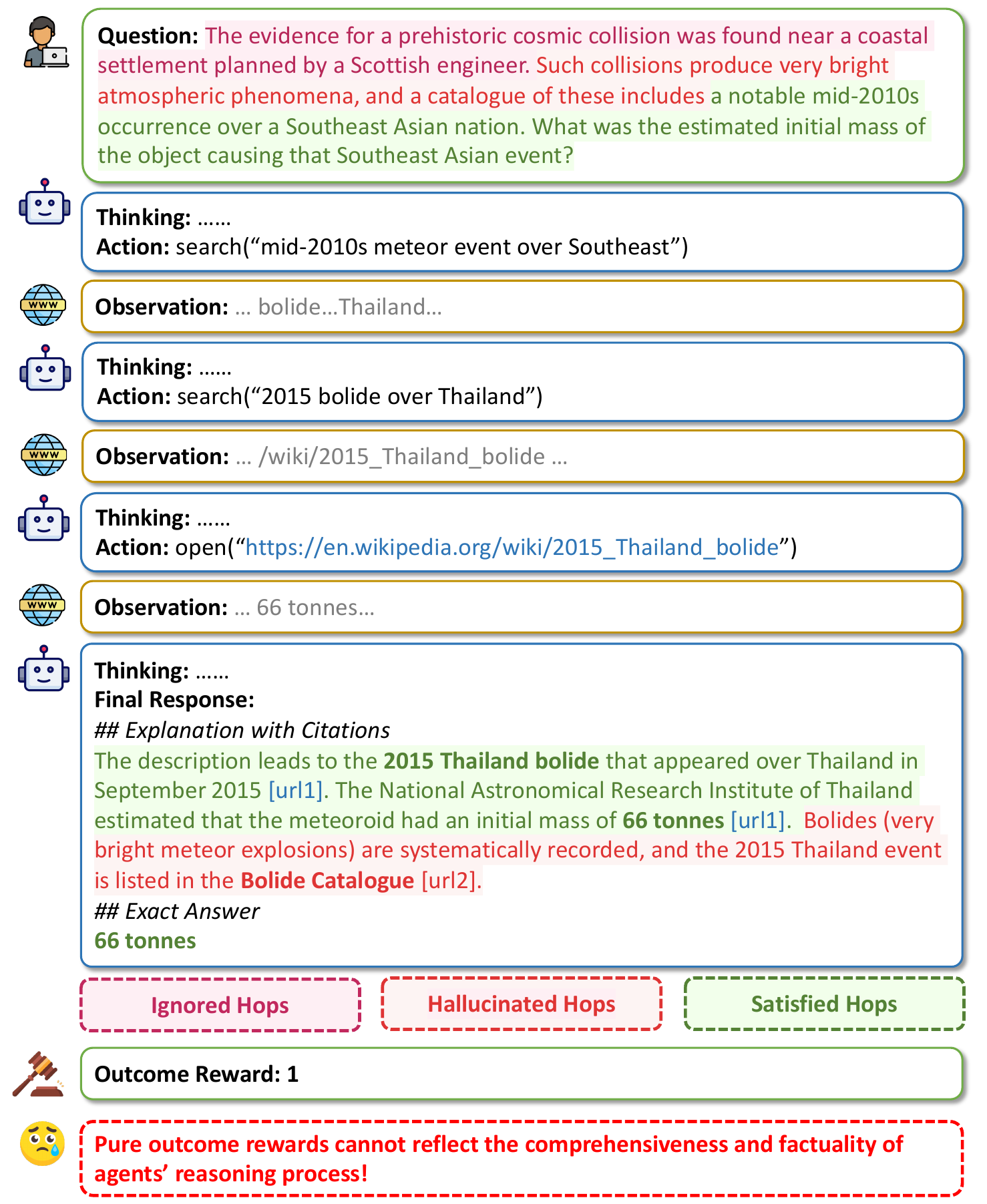}
    \caption{Pure outcome rewards fail to capture shortcut exploitation and hallucinations of deep search agents.}
    \label{fig:intro}
\end{figure}

\section{Introduction}
Recently, LLM-based deep search agents have attracted growing attention for their ability to leverage external web-browsing tools to solve complex, knowledge-intensive problems~\cite{yao2023react, wang2024survey, openai-deepresearch}. A prominent line of research has focused on applying reinforcement learning (RL) to further enhance these agents' long-horizon information-seeking capacity in the vast and noisy web environment, typically leveraging synthetic multi-hop QA datasets that are intentionally challenging but feature short-form answers for easy verification~\cite{gao2025asearcher, wu2025webdancer, li2025websailor, lu2025deepdive}. For the efficiency and scalability of RL, existing works commonly use only outcome rewards in training, which are binary signals indicating whether the agent's predicted final answer matches the ground truth~\cite{search-r1, gao2025asearcher, li2025websailor, liu2025webexplore}. 

While these outcome-based RL methods have demonstrated notable gains~\cite{li2025websailorv2, tongyidr}, they suffer from inherent limitations.
As illustrated in Figure~\ref{fig:intro}, binary outcome rewards alone cannot accurately reflect the comprehensiveness and factuality of agents' reasoning processes~\cite{shao2025deepseekmathv2}, leaving room for undesirable behaviours: Agents may arrive at the correct answer by shortcut solutions (e.g., exploiting only a few hops of information while ignoring other constraints in the question) or fortunate hallucination. Optimizing toward these flawed trajectories will result in deep search agents with diminished robustness and suboptimal performance. 

To address these limitations, we propose \textbf{Citation-aware Rubric Rewards (CaRR)}, a novel fine-grained reward framework for deep search agents that emphasizes reasoning comprehensiveness, factual grounding, and evidence connectivity. 
Our framework is inspired by the observation that each hop within the synthetic complex question can naturally serve as a checkpoint for evaluating the agent's reasoning process: An ideal trajectory that completely solves the given question should satisfy all hops by revealing the identities of all intermediate hidden entities and supporting them with correct citations. Building upon this idea, our framework first employs an LLM to decompose the multi-hop question into a list of single-hop factual statements, each involves several hidden entities that should be found during exploration. These factual statements are then used as point-wise rubrics to assess the comprehensiveness and factuality of agents' trajectories. Specifically, a rubric is satisfied by a trajectory only if (1) the identities of all relevant hidden entities are explicitly revealed in the final response; (2) the factual statement, along with the identified entities, is fully supported by the cited web contents; (3) the supported rubric can be connected to the predicted final answer via other supported rubrics, thereby constituting a complete evidence chain.  
Given a trajectory, we employ a judge LLM to check whether each rubric is satisfied following the above three criteria, and the citation-aware rubric reward is defined as the ratio of satisfied rubrics.

Building on CaRR, we further introduce \textbf{Citation-aware Group Relative Policy Optimization (C-GRPO)}, an extension of GRPO~\cite{shao2024deepseekmath} that incorporates context-aware rubric rewards with traditional outcome rewards in RL. Specifically, C-GRPO assigns an additional weighted rubric reward to the trajectories whose outcome reward is 1.
By doing so, C-GRPO preserves the primary objective of finding the correct answer while encouraging the agent to produce more comprehensive and evidence-grounded reasoning processes, thereby achieving robust RL and better final performance.  

To validate the efficacy of CaRR and C-GRPO, we conduct RL experiments on both small (4B) and large (30B) model scales. The evaluation results on four challenging deep search benchmarks indicate that C-GRPO consistently outperforms the GRPO baseline that uses pure outcome rewards, and also demonstrates significantly better performance when provided with extended context budgets. Our analysis reveals that C-GRPO successfully discourages shortcut exploitation and promotes more comprehensive, citation-supported solutions, yielding robust policies featured by rigorous self-verification and better factuality. Moreover, the agents trained with C-GRPO and synthetic QA data also generalize well on open-ended deep research tasks, even surpassing some advanced agents trained with proprietary data. 

In summary, our main contributions include: (1) We identify key limitations of outcome-based RL in training deep search agents, including shortcut exploitation and hallucination tolerance; (2) We propose CaRR, a novel framework that provides fine-grained rewards for assessing the comprehensiveness and factuality of deep search agents; (3) We propose C-GRPO, a mixed-reward RL algorithm combining outcome rewards and context-aware rubric rewards for training robust deep search agents; 
(4) We conduct extensive experiments and thorough analysis to validate the efficacy of CaRR and C-GRPO.

\begin{figure*}
    \centering
    \includegraphics[width=\linewidth]{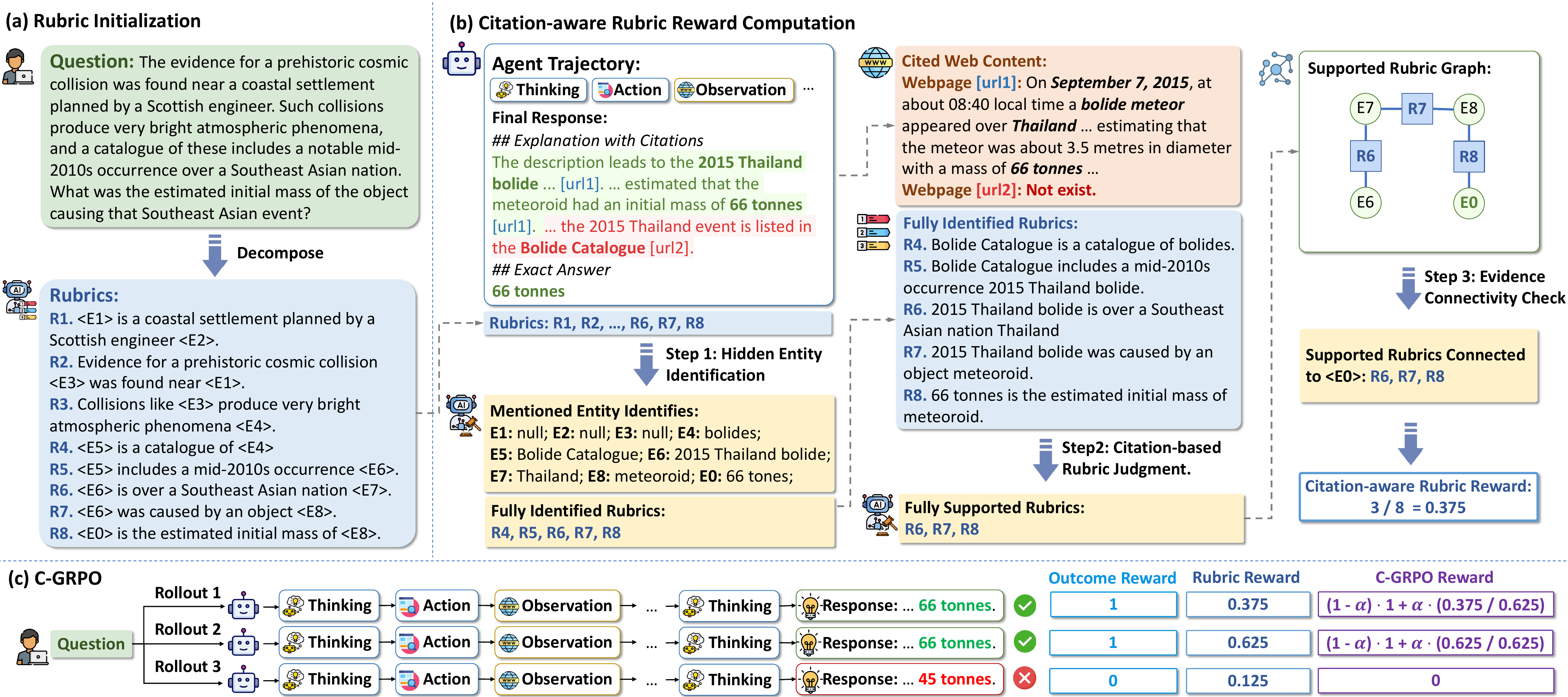}
    \caption{Overview of (a) rubric initialization; (b) computation of context-aware rubric rewards; (c) C-GRPO.}
    \label{fig:main}
\end{figure*}

\section{Methodology}
\label{sec:method}
In this section, we first provide a brief overview of key concepts in deep search agents, then introduce our CaRR framework and C-GRPO algorithm.
\subsection{Preliminary}
\paragraph{Deep Search Agents.}
We adopt the ReAct~\cite{yao2023react} paradigm for deep search agents. Given a question, the LLM-based agent follows an iterative cycle of thinking, action (i.e., a tool call), and observation until obtaining the final answer. A complete trajectory with $T$ iterations can be formalized as:
\begin{align}
    \mathcal{H} = (\tau_1, a_1, o_1, \dots, \tau_t, a_t, o_t, \dots, \tau_T, a_T),
\end{align}
where $\tau_t$, $a_t$, $o_t$ denote the thought, action, and observation at step $t$. Specifically, the action $a_t$ ($1 \le t < T$) calls one of the following three browsing tools: (1) a \texttt{search} tool that retrieves top-$n$ relevant webpages for the given query and returns the title, URL, and snippet of each webpage; (2) an \texttt{open} tool that accesses the given URL and shows the head part of the page; (3) a \texttt{find} tool that matches the given keyword in the opened webpage and returns surrounding content of each match. While $a_T$ is the final response, consisting of an explanation with citations and the final answer. The tool descriptions and trajectory format are detailed in Appendix~\ref{sec:traj_format}.

\paragraph{Synthetic Deep Search Training Data.}
RL of deep search agents typically relies on synthetic complex QA datasets~\cite{gao2025asearcher, li2025websailor, lu2025deepdive}. These datasets are commonly constructed from entity-centric knowledge graphs, involving multi-hop reasoning paths and deliberate information obfuscation to increase search complexity.
For training convenience, the final answer is often a short-form entity string, allowing automatic correctness verification.

\subsection{Citation-Aware Rubric Rewards}
To address the limitations of outcome rewards, we propose \textbf{Citation-aware Rubric Rewards (CaRR)}, a novel fine-grained reward framework for deep search agents, taking into account reasoning comprehensiveness, factual grounding, and evidence connectivity. Specifically, CaRR utilizes the underlying compositional structure of the synthetic data. As illustrated in Figure~\ref{fig:main}, CaRR first decomposes a synthetic multi-hop question into a list of atomic factual statements, each involves several hidden entities that need to be found. These atomic statements can naturally serve as point-wise, verifiable rubrics for assessing reasoning comprehensiveness and factuality of deep search agents: 
An ideal trajectory that completely solves the question should satisfy all rubrics by revealing the identities of corresponding hidden entities, supporting them with cited web contents, and connecting the supported rubrics to form complete evidence chains that link to the final answer. Moreover, the identified hidden entities and cited URLs should be detailed in the final response provided to the user.
Based on this idea, CaRR uses a three-step method after the rubric initialization to provide fine-grained reward for agent rollouts, including: (1) hidden entity identification; (2) citation-based rubric judgment; and (3) evidence connectivity check. We will detail the rubric initialization and the three-step reward computation as follows.

\subsubsection{Rubric Initialization}
For each question $q$ in the training set, we prompt an LLM $\mathcal{M}_\text{rubric}$ to decompose the question to locate hidden entities $\mathcal{E}_q$ (i.e., entities that should be found when solving $q$) and generate the initial rubrics $\mathcal{R}_q$:
\begin{align}
    \mathcal{E}_q, \mathcal{R}_q = \mathcal{M}_\text{rubric}(q),
\end{align}
where 
\begin{align}
    \mathcal{E}_q &= \{e_0, e_1, \dots, e_{n_q}\}, \notag \\
    \mathcal{R}_q &= \{r_1, \dots, r_{m_q}\}.
\end{align}
As illustrated in Figure~\ref{fig:main}, each hidden entity $e_i \in \mathcal{E}_q$ is denoted by a placeholder \texttt{<E$i$>}, and $e_0$ refers to the final answer. Each rubric $r_j = (s_j, \mathcal{E}_{q,j}) \in \mathcal{R}_q$ is an atomic factual statement $s_j$ about an entity set $\mathcal{E}_{q, j}\subseteq \mathcal{E}_q$, and will serve as a checkpoint for assessing search agent's trajectories. Note that these rubrics are pre-generated before training and remain unchanged throughout the RL process.

\subsubsection{Reward Computation}
\label{sec:three-step-reward}
After initializing the hidden entity set $\mathcal{E}_q$ and inital rubrics $\mathcal{R}_q$ for a question $q$, given an agent trajectory $\mathcal{H}=(\tau_1, a_1, o_1, \dots, \tau_T, a_T)$, we use a three-step procedure with a judge LLM $\mathcal{M}_\text{judge}$ to assign a fine-grained rubric reward for $\mathcal{H}$, taking into account reasoning comprehensiveness, citation grounding, and evidence connectivity:

\paragraph{Step 1: Hidden Entity Identification.}
From the perspective of reasoning comprehensiveness, 
an ideal trajectory for solving $q$ should consider all rubrics implied by $q$, uncover the identities of corresponding hidden entities during exploration, and explain them in the final response $a_T$. 
In light of this, we first employ $\mathcal{M}_\text{judge}$ to judge whether $a_T$ explicitly identifies the name of each hidden entity  $e_i \in \mathcal{E}_q$:
\begin{align}
    \{e_0^\mathcal{H}, \dots, e_n^\mathcal{H}\} = \mathcal{M}_\text{judge}(q, \mathcal{R}_q, \mathcal{E}_q, a_T),
\end{align}
where $e_i^\mathcal{H}$ is either the mentioned name of $e_i$ in $a_T$\footnote{Note that we do not require the identified $e_i^\mathcal{H}$ from $a_T$ to be equal to the golden entity $e_i^*$ used for constructing $q$.}, or \texttt{null} if the name is not explicitly identified. 
Only rubrics whose hidden entities are all identified are regarded as being fully identified by $\mathcal{H}$. Formally, by defining the mapping:
\begin{align}
    f^\mathcal{H}(e_i) = e_i^\mathcal{H}, \;\forall e_i\in \mathcal{E}_q\, 
\end{align}
we instantiate each $r_j = (s_j, \mathcal{E}_{q,j})\in \mathcal{R}_q$ by replacing hidden entities $\mathcal{E}_{q,j}$ with their identified name:
\begin{align}
    \mathcal{E}_{q, j}^\mathcal{H} &= \{f^\mathcal{H}(e_i) \mid e_i\in \mathcal{E}_{q,j} \}, \notag \\
    r_j^\mathcal{H} &= (s_j, \mathcal{E}_{q,j}^\mathcal{H}), \notag \\
    \mathcal{R}_q^\mathcal{H} &= \{r_1^\mathcal{H}, \dots, r_{m_q}^\mathcal{H}\}.
\end{align}
Then the fully-identified rubrics are defined as:
\begin{align}
    \mathcal{R}_q^\text{identify} &= \{r_j^\mathcal{H}\in \mathcal{R}_q^\mathcal{H} \mid e_i^\mathcal{H} \!\ne\! \texttt{null}, \forall e_i^\mathcal{H} \!\in\! \mathcal{E}_{q,j}^\mathcal{H} \},
\end{align}
which will be selected for further judgment.

\paragraph{Step 2: Citation-based Rubric Judgment.}
For each fully-identified rubric $r_j^\mathcal{H}\in \mathcal{R}_q^\text{identify}$, we further check whether $r_j^\mathcal{H}$ is grounded on the cited web contents in $\mathcal{H}$, preventing the agent from fabricating entity names or facts.
To achieve this, we first extract cited URLs\footnote{We extract at most 20 cited URLs to prevent the agent from hacking the reward by citing a large amount of webpages.} from the final response $a_T$ using regex
and collect corresponding web contents from $\mathcal{H}$ to form the supporting context $\mathcal{C}^\mathcal{H}$:
\begin{align}
    &url_1, \dots, url_k = \operatorname{ExtractCitation}(a_T), \notag \\
    &\mathcal{C}^\mathcal{H} = \operatorname{CollectContent}(\mathcal{H}, url_1, \dots, url_k), 
\end{align}
which includes deduplicated search snippets, opened webpage content, and keyword matches.  
Then we prompt the LLM $\mathcal{M}_\text{judge}$ to judge whether each identified rubric is fully supported by $\mathcal{C}^\mathcal{H}$:
\begin{align}
    \{s&p_1, \dots, sp_{m_q}\} = \mathcal{M}_\text{judge}(\mathcal{R}_q^\text{identify}, \mathcal{C}^\mathcal{H}), \notag \\
    &\mathcal{R}_q^\text{support} = \{r_j^\mathcal{H}\in \mathcal{R}_q^\text{identify} \mid sp_j = 1\},
\end{align}
where $sp_j\!\in\!\{0,1\}$ indicates whether $r_j^\mathcal{H}$ is supported. 

\paragraph{Step 3: Evidence Connectivity Check.}
Beyond individual support, we require that supported rubrics form connected evidence chains linked to the predicted answer entity $e_0^\mathcal{H}$. This prevents the agent from hacking a rubric by finding entities that satisfy the factual statement but are unrelated to $e_0^\mathcal{H}$. 
Specifically, we construct a bipartite graph:
\begin{align}
    \mathcal{G}^\mathcal{H} =\{\mathcal{E}_q^\mathcal{H}\cup \mathcal{R}_q^\text{support}, E\},
\end{align}
whose nodes are identified entities $\mathcal{E}_q^\mathcal{H}$ and supported rubrics $\mathcal{R}_q^\text{support}$, with an edge $(e_i^\mathcal{H}, r_j^\mathcal{H})\in E$ if $e_i^\mathcal{H}$ appears in $r_j^\mathcal{H}$, i.e., $e_i^\mathcal{H} \in \mathcal{E}_{q,j}^\mathcal{H}$.  
Then we apply a breadth-first search (BFS) starting from $e_0^\mathcal{H}$ to determine the set of reachable rubrics \( \mathcal{R}^\text{connect} \):
\begin{align}
    \mathcal{R}^\text{connect} = \{ r_j^\mathcal{H} \mid r_j^\mathcal{H}\text{ is connected to }e_0^\mathcal{H}\text{ in }\mathcal{G}^\mathcal{H}\} 
\end{align}
The final rubric reward is given by:
\begin{align}
    R^\mathcal{H}_\text{r} = \frac{|\mathcal{R}_q^\text{connect}|}{|\mathcal{R}_q|},
\end{align}
which measures the proportion of rubrics that are fully identified, citation-supported, and logically connected to the predicted answer.

\subsection{C-GRPO}
Based on CaRR framework, we further introduce \textbf{Citation-aware Group Relative Policy Optimization (C-GRPO)}, which combines citation-aware rubric rewards and outcome rewards in GRPO for training robust deep search agents. Specifically, C-GRPO assigns an additional weighted rubric reward to the trajectories whose outcome reward is 1. By doing so, C-GRPO preserves the primary objective of finding the correct answer while encouraging the agent to produce more comprehensive and evidence-grounded reasoning processes. Formally, let $\mathcal{H}_1, \dots, \mathcal{H}_G$ be a group of rollout for a question $q$, whose ground truth answer is $gt$. We first obtain the outcome reward $R^{\mathcal{H}_i}_\text{o}$ (i.e., whether $\mathcal{H}_i$ finds $gt$) and the context-aware rubric reward $R^{\mathcal{H}_i}_\text{r}$ for each $\mathcal{H}_i$. Then the mixed reward of $\mathcal{H}_i$ is defined as:
\begin{align}
    R_i = (1 - \alpha) \cdot R^{\mathcal{H}_i}_\text{o} + \alpha \cdot R^{\mathcal{H}_i}_\text{o} \cdot \hat{R}^{\mathcal{H}_i}_\text{r},
\label{eq:reward}
\end{align}
where $\alpha\in [0,1]$ is a hyperparameter balancing outcome and rubric rewards, and  we use the normalized rubric rewards
\begin{align}
    \hat{R}^{\mathcal{H}_i}_\text{r} = \frac{R^{\mathcal{H}_i}_\text{r}}{\max_{j\in\{1, \dots, G\}} R^{\mathcal{H}_j}_\text{r}}
\end{align}
to stable advantage calculation across different groups. In addition, rollouts with format error or overlength problem (i.e., exceeding token or tool-call limits) are assigned a reward of 0. 
Finally, the agent policy is optimized by maximizing a multi-turn GRPO objective with token-level loss:
\begin{align}
&\mathcal{J}(\theta)
= \mathbb{E}_{%
\substack{
(q, gt) \sim \mathcal{D},\\
\{\mathcal{H}_i\}_{i=1}^G \sim \pi_{\theta_{\text{old}}}(\cdot \mid q)
}
}
\bigg[
{\textstyle\frac{\text{\normalsize $1$}}{\sum\limits_{i=1}^G\sum\limits_{j=1}^{|\mathcal{H}_i|}I(\mathcal{H}_{i,j})}
\sum\limits_{i=1}^G \sum\limits_{j=1}^{|\mathcal{H}_i|}} \notag \\
& I(\mathcal{H}_{i,j})
\min\big( 
\rho_{i,j} \hat{A}_{i,j}, 
\operatorname{clip}(
\rho_{i,j})_{1\!-\!\epsilon_{\text{low}}}^{1\!+\!\epsilon_{\text{high}}}
\hat{A}_{i,j}
\big) 
\bigg].
\end{align}
where $\mathcal{H}_{i,j}$ denotes the $j$-th token of $\mathcal{H}_i$; $\rho_{i,j} = \frac{\pi_\theta(\mathcal{H}_{i,j}|q, \mathcal{H}_{i,1:j-1})}{\pi_{\theta_\text{old}}(\mathcal{H}_{i,j}|q, \mathcal{H}_{i,1:j-1})}$ and $\hat{A}_{i,j}=\frac{R_i-\operatorname{mean}(R_k)_{k=1}^G}{\operatorname{std}(R_k)_{k=1}^G}$ are the importance sampling ratio and advantage of $\mathcal{H}_{i,j}$; and $I(\mathcal{H}_{i,j})\in \{0,1\}$ indicates whether $\mathcal{H}_{i,j}$ is generated by the LLM itself (i.e., not from observed web content).

\begin{table*}[!t]
\centering
\resizebox{\linewidth}{!}{
\begin{tabular}{l|c|C{3cm}|C{3cm}|C{3cm}|C{3cm}}
\toprule
\textbf{Model}         & \textbf{Context Budget} & \textbf{BrowseComp}  & \textbf{BrowseComp-ZH} & \textbf{xbench-DS} & \textbf{GAIA}        \\ \midrule
\multicolumn{6}{c}{\cellcolor[HTML]{D8D8D8}\textit{\textbf{Advance Agents with Proprietary Data}}}                                                   \\ \midrule
OpenAI-o3              & 128k                    & 49.7                 & 58.1                   & 66.7                       & 70.5                 \\
DeepSeek-v3.1          & 128k                    & 30.0                 & 49.2                   & 71.2                       & 63.1                 \\
Tongyi-DeepResearch    & 128k                    & 43.4                 & 46.7                   & 75.0                       & 70.9                 \\
GLM-4.5                & 128k                    & 26.4                 & 37.5                   & 70.0                       & 66.0                 \\
GLM-4.6                & 128k                    & 45.1                 & 49.5                      & -                          & -                    \\ \midrule
\multicolumn{6}{c}{\cellcolor[HTML]{D8D8D8}\textit{\textbf{Agents with Open-source Data}}}                                                           \\ \midrule
Asearcher-Web-32B      & 128k                    & 5.2                  & 15.6                   & 42.1                       & 52.8                 \\
WebSailor-7B           & 32k                     & 6.7                  & 14.2                   & 34.3                       & 37.9                 \\
WebSailor-32B          & 32k                     & 10.5                 & 25.5                   & 53.3                       & 51.5                 \\
WebExplorer-8B         & 128k                    & 15.2                 & 32.0                   & 53.7                       & 50.0                 \\
DeepDive-9B            & 128k                    & 6.3                  & 15.1                   & 51.8                       & -                    \\
DeepDive-32B           & 128k                    & 15.3                 & 29.7                   & 51.8                       & -                    \\ \midrule
\multicolumn{6}{c}{\cellcolor[HTML]{D8D8D8}\textit{\textbf{Our Agents}}}                                                                             \\ \midrule
DeepDive-4B-SFT  & 64k / 128k                & 7.7 / 14.1           & 10.1 / 16.6            & 34.0 / 44.3                & 39.5 / 46.0          \\
+ GRPO                 & 64k / 128k                & 12.9 / 14.7          & 16.6 / 17.5            & 41.0 / 41.3                & 40.5 / 41.1          \\
+ E-GRPO               & 64k / 128k                & 11.5 / 14.5          & 16.5 / 20.2            & 43.7 / 45.0                & 42.4 / 42.4          \\
+ C-GRPO (ours)        & 64k / 128k                & \textbf{13.9 / 17.5} & \textbf{18.2 / 24.7}   & \textbf{50.3 / 54.0}       & \textbf{48.9 / 50.2} \\ \midrule
DeepDive-30B-SFT & 64k / 128k                & 12.2 / 20.5          & 15.8 / 24.7            & 43.0 / 54.3                & 46.0 / 50.8          \\
+ GRPO                 & 64k / 128k                & 16.0  / 18.9         & 24.1 / 26.1            & 51.3 / 52.0                & 51.1 / 51.1          \\
+ E-GRPO               & 64k / 128k                & 13.1 / 18.5          & 17.1 / 24.0            & 51.7 / 55.7                & 52.8 / 55.3          \\
+ C-GRPO (ours)        & 64k / 128k                & \textbf{17.9 / 24.8} & \textbf{26.0 / 33.3}   & \textbf{55.3 / 57.7}       & \textbf{53.7 / 56.3} \\
\bottomrule
\end{tabular}
}
\setlength{\abovecaptionskip}{0.1cm}
\caption{Overall performance of different agents on four challenging deep search benchmarks.}

\label{tab:main}
\end{table*}

\begin{figure*}
    \centering
    \includegraphics[width=\linewidth]{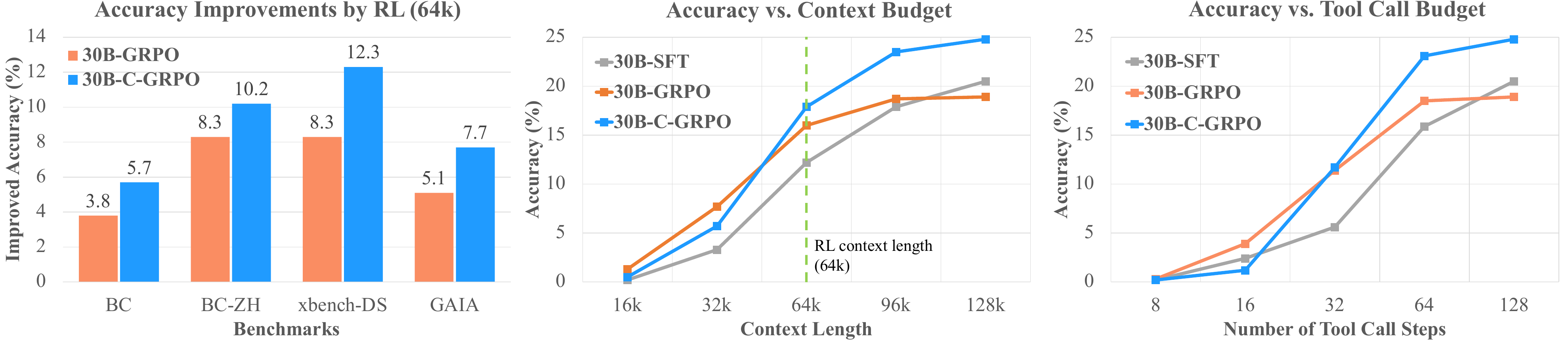}
    \caption{\textit{Left}: Accuracy improvements by GRPO and C-GRPO over SFT models at 64k context length. \textit{Middle} and \textit{Right}: Test-time scaling performance of different models with respect to context budget and tool call budget.}
    \label{fig:test-time scaling}
\end{figure*}

\begin{figure*}
    \centering
    \includegraphics[width=\linewidth]{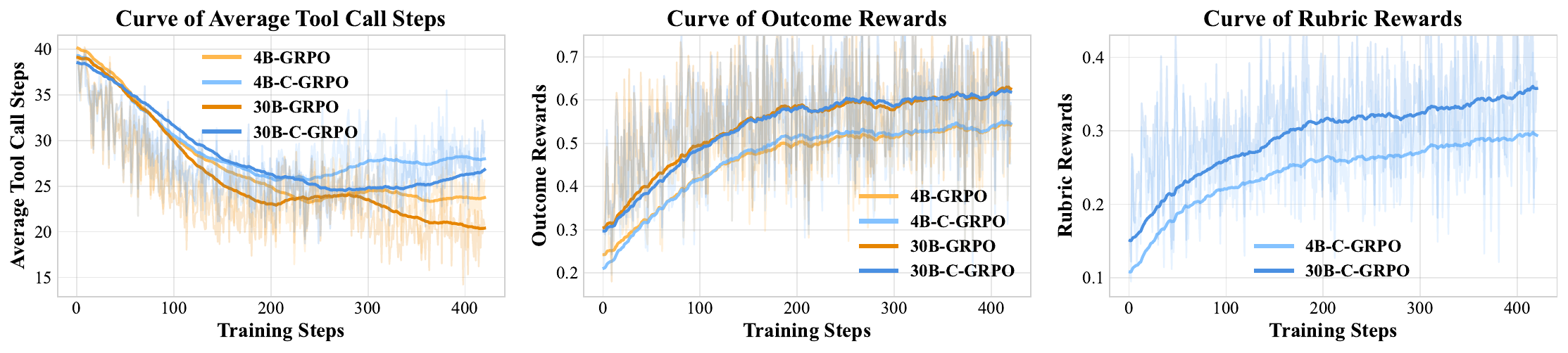}
    \caption{Training dynamics of GRPO and C-GRPO, including the changes of average tool call steps, outcome rewards, and rubric rewards.}
    \label{fig:train_dynamics}
\end{figure*}

\section{Experiments}
In this section, we conduct RL experiments to show the effectiveness of context-aware rubric rewards and C-GRPO for training deep search agents.

\subsection{Experiment Setup} 
\paragraph{Models and Training Data.} We select Qwen3-4B-Thinking-2507~\cite{qwen3} and Qwen3-30B-A3B-Thinking-2507~\cite{qwen3} as our backbone models, covering different model sizes and architectures (dense and MoE). We use DeepDive~\cite{lu2025deepdive}, an open-sourced deep search dataset, as our training data. This dataset is automatically synthesized through knowledge graph random walks and entity obfuscation, consisting of 1,016 samples for SFT and 2,234 samples for RL.

\paragraph{Environment Settings} We use Serper API~\cite{serper} for the \texttt{search} tool. The \texttt{open} tool first fetches the webpage using Jina API~\cite{jina}, and then returns the first 10k chars. The \texttt{find} tool is implemented with vanilla string matching.

\paragraph{Baselines.} To demonstrate the algorithm enhancement, we compare C-GRPO with two baseline RL algorithms for deep search agents: (1) GRPO~\cite{shao2024deepseekmath, search-r1}, which only uses the outcome rewards and is widely adopted in previous works; (2) E-GRPO~\cite{zhao2025egrpo}, which takes normalized entity match rate (i.e., the ratio of golden hidden entities identified during an agent’s reasoning process), as the fine-grained rewards for incorrect rollouts to distinguish ``near-miss'' samples from complete failures. Besides, we present the reported scores of several state-of-the-art search agents as references, though they may adopt different training data and context lengths from us. We detail them in Appendix~\ref{sec:baselines}.

\paragraph{Training Details.} Our training process includes cold-start SFT and subsequent RL. For cold-start SFT, we first leverage GLM-4.6~\cite{glm45} to generate 832 high-quality SFT traces through reject sampling on the SFT split of DeepDive dataset. Then we train each model on these traces for 3 epochs with a batch size of 16, a learning rate of 4e-5, and a maximum context length of 128k. For RL, we use all 2,234 QA pairs from the DeepDive RL split. The training configuration includes a rollout size of 16, 8 samples per prompt, a global batch size of 128, a temperature of 1.0, a learning rate of 2e-6, and a maximum context length of 64k tokens. We train each model for 3 epochs. We set the rubric reward weight $\alpha$ to be 0.3, and the effect of different values of $\alpha$ can be found in Sec.. We use DeepSeek-v3.2~\cite{deepseek32} as the judge LLM for both outcome rewards and rubric rewards.

\paragraph{Benchmarks and Evaluation Details.} We evaluate the trained agents on four challenging deep search benchmarks, including: BrowseComp~\cite{wei2025browsecomp}, BrowseComp-ZH~\cite{zhou2025browsecompzh}, xbench-DeepSearch~\cite{xbench2025}, and the text-only validation subset of GAIA~\cite{mialon2024gaia}. 
These benchmarks comprehensively assess the essential capabilities for effective deep search in long-horizon information-seeking, multi-step web navigation, complex reasoning, and cross-lingual synthesis. 
Following their official LLM-as-judge settings, we use GPT-5-Chat~\cite{gpt-5} to assess whether the agent’s final output matches the ground truth answer. Considering the relatively small dataset size of BrowseComp-ZH, xbench-DeepSearch, and GAIA, we repeat their evaluation 3 times and report the average accuracy. In addition, we evaluate each model at both 64k and 128k context lengths, where the former corresponds to the context length of RL, and the latter is used to assess the test-time scaling capacities given abundant context budgets.


\subsection{Main Result}
We present the main experimental results in Table~\ref{tab:main}. As shown in the table, our proposed C-GRPO significantly outperforms GRPO and E-GRPO baselines on all benchmarks across both 4B and 30B scales. Specifically, with the 64k/128k context budget, C-GRPO achieves an average improvement of 5.1/8.0 for the 4B model and 2.6/6.0 for the 30B model compared to GRPO. Surprisingly, we find that though GRPO with pure outcome reward notably improves the performance of SFT models within the RL context length (i.e., 64k), it may compromise their test-time scaling performance on longer context length (i.e., 128k). Our training dynamic analysis and case studies indicate that this compromise stems from the inherent limitation of pure outcome rewards, which leaves room for shortcut exploitation and hallucinations. In contrast, C-GRPO consistently improves the SFT models at both 64k and 128k context lengths, demonstrating the effectiveness of context-aware rubric rewards for training more robust deep search agents. Moreover, our trained models with C-GRPO achieve state-of-the-art performance among agents using open-source data, narrowing the gap with advanced agents that use proprietary data.

\begin{table}[!t]
\centering
\resizebox{\linewidth}{!}{
    \setlength{\tabcolsep}{3pt}
\begin{tabular}{l|ccccc}
\toprule
\textbf{Model}   & $|\mathcal{C}^\mathcal{H}|$ & $|\mathcal{R}_q^\text{identify}|$ & $|\mathcal{R}_q^\text{support}|$ & $|\mathcal{R}_q^\text{connect}|$ & $|\mathcal{R}_q|$ \\ \midrule
\toprule
DeepDive-30B-SFT & 3.8                         & 8.0                               & 6.2                              & 4.5                              & 10.1              \\
+ GRPO           & 3.5                         & 7.5                               & 5.3                              & 4.0                              & 10.1              \\
+ C-GRPO (ours)  & \textbf{4.3}                & \textbf{8.2}                      & \textbf{6.6}                     & \textbf{5.2}                     & 10.1       \\
\bottomrule
\end{tabular}
}
\setlength{\abovecaptionskip}{0.1cm}
\caption{Comparison of the number of cited webpages and rubric satisfaction on a subset of BrowseComp.}

\label{tab:browsecomp_rubrics}
\end{table}

\subsection{More Analysis}
\label{sec:more_analysis}

\paragraph{Traning dynamics.} 
We present the training dynamics of GRPO and C-GRPO in Figure~\ref{fig:train_dynamics}. 
As illustrated, the average tool call steps of GRPO and C-GRPO both decline at the beginning of training, where the agents learn to improve their search efficiency to avoid overlength rollouts. 
As the training progresses, the tool call steps of GRPO continue to decrease after a slight increase, implying that the models fall into a local optimal policy that favors shortcut solutions. 
Case studies in Appendix~\ref{sec:case_study} show that the GRPO agent becomes prone to finding an answer based on the last few hops of the question without thoroughly verifying other constraints. 
While such a policy can yield high outcome rewards within limited context budgets, it sacrifices performance on more difficult questions that require careful verification using longer contexts. 
Moreover, outcome rewards alone are insufficient to guide agents out of this local optimum, as they cannot punish the shortcut exploitation behaviors.
In contrast, the tool call steps of our C-GRPO keep increasing after the initial decline, suggesting that the models are trying to satisfy more rubrics by gathering more evidence to support and verify their predicted answer, which results in a more robust policy. 
During the same period, the outcome rewards of C-GRPO even slightly exceed GRPO, further validating that our mixed fine-grained rewards provide a more effective and robust learning signal than pure outcome rewards.

\paragraph{Comprehensiveness and factuality.} 
To assess the impact of different RL algorithms on the comprehensiveness and factuality of agents, we analyze the rubric satisfaction of 30B agents on the evaluation sets. Specifically, we generate rubrics for a subset of BrowseComp in which all agents solve the queries within a 64k context length, and compare the number of their cited webpages and satisfied rubrics. As shown in Table~\ref{tab:browsecomp_rubrics}, the C-GRPO agent cites more webpages and satisfies more rubrics than the SFT and GRPO baselines, indicating that C-GRPO effectively enhances agent comprehensiveness and factuality by incentivizing more extensive evidence gathering. Conversely, the cited webpages and satisfied rubrics of the GRPO agent are both fewer than the SFT baseline, further validating the shortcut exploitation issue of pure outcome rewards. 

\begin{table}[!t]
\centering
\resizebox{\linewidth}{!}{
    \setlength{\tabcolsep}{4pt}
\begin{tabular}{l|ccccc}
\toprule
                                 & \multicolumn{5}{c}{\textbf{DeepReasearch Bench}}                                                                \\ \cmidrule(lr){2-6}
\multirow{-2}{*}{\textbf{Model}} & \textbf{Overall} & \textbf{Comp.} & \textbf{Insight} & \textbf{Inst.} & \textbf{Read.} \\ \midrule
\multicolumn{6}{c}{\cellcolor[HTML]{D8D8D8}\textit{\textbf{Advance Agents with Proprietary Data}}}                                                \\ \midrule
OpenAI-DeepResearch              & 46.45            & 46.46                      & 43.73            & 49.39                & 47.22                \\
Kimi-Researcher                  & 44.64            & 44.96                      & 41.97            & 47.14                & 45.59                \\
Tongyi-DeepResearch              & 40.46            & 39.46                      & 34.44            & 46.22                & 44.27                \\
Grok-Deeper-Search               & 38.22            & 36.08                      & 30.89            & 46.59                & 42.17                \\ \midrule
\multicolumn{6}{c}{\cellcolor[HTML]{D8D8D8}\textit{\textbf{Our Agents}}}                                                                          \\ \midrule
DeepDive-4B-SFT            & 33.81            & 29.57                      & 24.23            & 44.05                & 41.02                \\
+ GRPO                           & 34.79            & 31.29                      & 26.79            & 43.81                & 41.58                \\
+ E-GRPO                         & 36.59            & 33.20                      & 28.30            & 45.58                & 42.67                \\
+ C-GRPO (ours)                  & \textbf{37.51}   & \textbf{33.88}             & \textbf{30.01}   & \textbf{45.72}       & \textbf{43.82}       \\ \midrule
DeepDive-30B-SFT           & 37.51            & 34.27                      & 28.85            & 46.77                & 43.21                \\
+ GRPO                           & 39.30            & 36.10                      & 31.66            & 47.65                & 44.92                \\
+ E-GRPO                         & 36.12            & 32.31                      & 27.73            & 45.72                & 42.33                \\
+ C-GRPO (ours)                  & \textbf{41.99}   & \textbf{39.75}             & \textbf{35.87}   & \textbf{48.51}       & \textbf{46.63} \\
\bottomrule
\end{tabular}
}
\setlength{\abovecaptionskip}{0.1cm}
\caption{Performance of different agents on DeepResearch Bench across four dimensions, including comprehensiveness (Comp.), insight, instruction following (Inst.), and readability (Read.).}
\label{tab:deepresearchbench}
\end{table}

\paragraph{Generalize to open-ended deep research tasks.}
To assess the generalization capabilities of agents trained using C-GRPO and synthetic data in open-ended deep research tasks, we conduct evaluations on DeepResearch Bench~\cite{jin2025deepresearchbench}, where the agents are required to write research reports for PhD-level tasks, 
and the generated reports are assessed by Gemini-2.5-Pro-preview~\cite{gemini-25-pro-preview} based on the pre-defined rubrics spanning multiple dimensions. 
As shown in Table~\ref{tab:deepresearchbench}, C-GRPO consistently surpasses other RL algorithms and yields substantial improvements over SFT models in all dimensions. Moreover, the 30B model trained with C-GRPO even outperforms several advanced agents using proprietary data, demonstrating the strong generalization abilities of our approach.

\begin{table}[!t]
\centering
\resizebox{\linewidth}{!}{
    \setlength{\tabcolsep}{8pt}
\begin{tabular}{l|cccc}
\toprule
\textbf{Model}          & \textbf{BC} & \textbf{BC-ZH} & \textbf{xbench-DS} & \textbf{GAIA} \\ \midrule
DeepDive-4B-SFT         & 14.1                & 16.6                   & 44.3               & 46.0          \\
+ C-GRPO ($\alpha$=0)   & 14.7                & 17.5                   & 41.3               & 41.1          \\
+ C-GRPO ($\alpha$=0.1) & 13.0                & 18.0                   & 46.0               & 46.3          \\
+ C-GRPO ($\alpha$=0.3) & \textbf{17.5}       & \textbf{24.7}          & \textbf{54.0}      & \textbf{50.2} \\
+ C-GRPO ($\alpha$=0.5) & 17.0                & 20.8                   & 49.3               & 42.4          \\
\bottomrule
\end{tabular}
}
\setlength{\abovecaptionskip}{0.1cm}
\caption{Performance of C-GRPO with different $\alpha$.}
\label{tab:effect_of_alpha}
\end{table}
\begin{table}[!t]
\centering
\resizebox{\linewidth}{!}{
    \setlength{\tabcolsep}{1.5pt}
\begin{tabular}{l|cccc}
\toprule
\textbf{Model}                          & \textbf{BC} & \textbf{BC-ZH} & \textbf{xbench-DS} & \textbf{GAIA} \\ \midrule
DeepDive-4B-C-GRPO                      & \textbf{17.5}       & \textbf{24.7}          & \textbf{54.0}      & \textbf{50.2} \\
w/o Hidden entity identification        & 16.5                & 23.2                   & 50.7               & 46.6          \\
w/o Evidence connectivity check         & 15.1                & 20.8                   & 47.7               & 44.0          \\
w/ Rubric rewards for all rollouts & 13.3                & 14.0                   & 40.3               & 40.8          \\
\bottomrule
\end{tabular}
}
\setlength{\abovecaptionskip}{0.1cm}
\caption{Performance of C-GRPO (1) without hidden entity identification; (2) without evidence connect check; (3) that adds weighted rubric rewards for all rollouts.}

\label{tab:ablation}
\end{table}

\subsection{Ablation Studies}
In this section, we conduct ablation studies using the 4B model to demonstrate the effect of each component in the CaRR framework and C-GRPO.
\label{sec:ablation}
\paragraph{Effect of rubric reward weight.} 
To illustrate the effect of the rubric reward weight $\alpha$ in C-GRPO, we train the 4B model with different $\alpha$ values, ranging from 0 (which is just GRPO) to 0.5. As shown in Table~\ref{tab:effect_of_alpha}, the overall performance gradually improves as $\alpha$ increases from 0, peaking at 0.3. This demonstrates the benefit of incorporating context-aware rubric rewards in RL. However, the performance begins to decrease as $\alpha$ becomes larger, suggesting that the model is distracted from the primary goal of finding a correct final answer. Therefore, it is important to use a moderate $\alpha$ value to balance the two reward components to obtain the optimal policy. 

\paragraph{Effect of hidden entity identification.} 
To show the effect of hidden entity identification in CaRR, we remove this step and let the judge LLM directly select the supported rubrics based on the model response and cited web contents, without considering whether each rubric is fully identified. As shown in Table~\ref{tab:ablation}, this ablation leads to a clear performance drop for C‑GRPO, suggesting that enforcing a stricter reward process via hidden entity identification enhances the effectiveness of RL.

\paragraph{Effect of evidence connectivity check.} 
We show the effect of the evidence connectivity check in CaRR by eliminating this step and instead setting the rubric reward to the fraction of supported rubrics, i.e., $R_{r}^\mathcal{H}=\frac{\mathcal{R}_q^\text{support}}{|\mathcal{R}_q|}$. The results in Table \ref{tab:ablation} show a substantial decline in performance without the connectivity check, since the agents learns to hack rubrics by finding entities that satisfy isolated factual statements but are unrelated to the final answer.

\paragraph{Adding rubric rewards for all rollouts.}
According to Equation~\ref{eq:reward}, we only add weighted context-aware rubric rewards for correct rollouts whose outcome reward is 1. If we add the rubric rewards for all rollouts, the advantage of some incorrect rollouts will receive positive advantages when there are few correct rollouts or many overlength rollouts in a group, which frequently happens at the beginning of RL. As a result, the model will be incorrectly optimized and perform badly, as shown in Table~\ref{tab:ablation}. 
\section{Related Works}

\paragraph{RL for Deep Search Agents.} 
Recently, RL has emerged as a critical technique for enhancing deep search agents~\cite{openai-deepresearch, search-r1}. Existing works can be broadly divided into two categories. The first category focuses on complex QA data synthesis and infrastructure design to support RL training~\cite{gao2025asearcher, wu2025webdancer, li2025websailor, lu2025deepdive, liu2025webexplore, li2025websailorv2, tongyidr}, and the second category focuses on improving RL algorithms to better fit multi-turn agentic settings~\cite{feng2025gigpo, dong2025arpo, dong2025aepo}. Nonetheless, these works typically rely solely on outcome rewards, with limited attention devoted to addressing the their limitations. E-GRPO~\cite{zhao2025egrpo} proposes to use the entity match rate as the fine-grained rewards for incorrect rollouts to distinguish ``near-miss'' samples from complete failures. However, it relies on gold annotations for intermediate hidden entities, and we also observed that applying fine-grained rewards for incorrect rollouts may mislead the RL optimization (see Sec.~\ref{sec:ablation}). 

\paragraph{Aligning LLMs with Rubric Rewards.} 
Recently, a series of works have explored the use of rubrics~\cite{arora2025healthbench, asai2024openscholar} in aligning LLMs for complex instruction following~\cite{lambert2024tulu3, dong2025autoif, peng2025verif} and long-form generation tasks~\cite{gunjal2025rubrics, shao2025drtulu}, where traditional reward models fail to provide reliable supervision signals. Specifically, they equip each training instance with a list of verifiable rubrics, and the reward of a model response is given by the ratio of its satisfied rubrics. Some works also explore evolving rubrics during training by contrasting multiple model rollouts~\cite{shao2025drtulu, rezaei2025onlinerubrics, wu2025rlac}. In this work, we show that rubric rewards can be utilized to supervise agents' reasoning processes, serving as an effective auxiliary of traditional outcome rewards.

\section{Conclusion}
In this work, we propose CaRR, a novel framework that provides fine-grained rewards for deep search agents, taking into account reasoning comprehensiveness, factual grounding, and evidence connectivity. We further introduce C-GRPO, which combines CaRR and outcome rewards in RL for training robust deep search agents. Our extensive experiments demonstrate that C-GRPO achieves significant improvement over GRPO in both deep search benchmarks and open-ended research tasks.  

\section{Limitations}
As described in Sec.~\ref{sec:method}, our rubric generation relies on the compositional structure of synthetic multi-hop questions, and may not be able to be directly adapted to open-ended QA training where some requirements are not explicitly stated in the question. Nonetheless, both our work and previous works have shown that the synthetic, short-form question answering is an effective proxy for open-ended deep research tasks since they share the core requirement for long-horizon information-seeking capacity. Moreover, the improvement in reasoning comprehensiveness and factual grounding brought by our context-aware rubric rewards also benefits the model's performance on open-ended deep research tasks, as demonstrated in Sec.~\ref{sec:more_analysis}.

\section{Ethical Considerations}
All the models and datasets used in this work are publicly published with permissible licenses.

\bibliography{custom}

\clearpage
\appendix
\newcommand{\user}[1]{\textcolor{cyan}{\texttt{<|im\_start|>user\\}}#1 \textcolor{cyan}{\texttt{<|im\_end|>}}}
\newcommand{\tool}[1]{\textcolor{orange}{\texttt{<|im\_start|>user\\}}#1 \textcolor{orange}{\texttt{<|im\_end|>}}}
\newcommand{\assistant}[1]{\textcolor{blue}{\texttt{<|im\_start|>assistant\\}}#1 \textcolor{blue}{\texttt{<|im\_end|>}}}
\newcommand{\think}[1]{\textcolor{blue}{\texttt{<think>\\}}#1 \textcolor{blue}{\texttt{\\</think>\\}}}
\newcommand{\toolcall}[1]{\textcolor{blue}{\texttt{<tool\_call>\\}}#1\textcolor{blue}{\texttt{\\</tool\_call>}}}
\newcommand{\toolresponse}[1]{\textcolor{orange}{\texttt{<tool\_response>\\}}#1 \textcolor{orange}{\texttt{\\</tool\_response>}}}

\section{Trajectory Format}
\label{sec:traj_format}
We show our tool descriptions and trajectory format in Figure~\ref{fig:tool_desc} and~\ref{fig:trajectory_format}, respectively.

\section{Details of Referred Deep Search Agents}
\label{sec:baselines}
For deep search benchmarks, we present the scores of OpenAI o3~\cite{openai-o3}, DeepSeek-v3.1~\cite{deepseek-v31}, Tongyi-DeepResearch~\cite{tongyidr}, GLM-4.5~\cite{glm45}, GLM-4.6~\cite{glm45}, Aseacher~\cite{gao2025asearcher}, WebSailor~\cite{li2025websailor}, WebExplorer~\cite{liu2025webexplore}, and DeepDive~\cite{lu2025deepdive} from their official reports or previous papers. For DeepResearch Bench, we present the scores of OpenAI-DeepDeesearch~\cite{openai-deepresearch}, Kimi-Researcher~\cite{kimi-researcher}, Tongyi-DeepResearch~\cite{tongyidr}, and Grok-Deeper-Search~\cite{xai_deep_research} from the official leaderboard~\cite{jin2025deepresearchbench}.

\section{Human Verification for LLM Judge}
To assess the reliability of the judge LLM in identifying hidden entities and applying citation-based rubric evaluations within the CaRR framework, we conducted a manual review of its judgments across 10 DeepDive‑30B‑SFT trajectories, covering 128 hidden entities and 164 rubrics. Using human assessments as the gold standard, the judge LLM achieved accuracies of 97.7\% for hidden entity identification and 95.1\% for citation‑based rubric evaluation, indicating strong reliability.

\section{Case Studies}
\label{sec:case_study}
To highlight the qualitative differences of GRPO and C-GRPO, we compare trajectories produced by DeepDive‑30B‑GRPO and DeepDive‑30B‑C‑GRPO for the same queries in both the training set (DeepDive) and the evaluation set (BrowseComp). We present only the final turn of each trajectory, which is sufficient to demonstrate the key distinctions. As shown in Case 1 (Figure~\ref{case:train_bad}), the GRPO agent tends to infer the answer based on the last several hops in the question, without carefully checking other constraints. It often guesses the identities of entities in the head part of the question (marked in \textcolor{red}{red}) without further verification. Such an unrobust policy is prone to failure on more challenging questions that demand thorough validation, as illustrated in Case 4 (Figure\ref{case:eval_bad}). n contrast, the C‑GRPO agent, as indicated by the \textcolor{ForestGreen}{green} highlights in Case 2 (Figure\ref{case:train_good}) and Case 4 (Figure\ref{case:eval_good}), continues to gather evidence until it can confirm that every constraint in the query is satisfied. It further ensures that each statement in its response is supported by corresponding citations.

\section{Prompts}
We show our used prompts in Figure~\ref{prompt:outcome_reward},~\ref{prompt:rubric_initialization},~\ref{prompt:entity_identify}, and~\ref{prompt:rubric_judgement}.

\begin{figure*}[htbp]
  \centering
\begin{tcolorbox}[title=\textbf{Tool Descriptions}, fontupper=\small]
\textbf{browser.search}\\
Description: Search in browser\\
Parameters:
\begin{itemize}[nosep]
    \item query: string, Search query
    \item num: integer, Number of results to return (default 10)
\end{itemize}

\medskip

\textbf{browser.open}\\
Description: Open browser link\\
Parameters:
\begin{itemize}[nosep]
    \item id: integer or string, ID or URL of the link to open
\end{itemize}

\medskip

\textbf{browser.find}\\
Description: Find pattern in the opened browser content\\
Parameters:
\begin{itemize}[nosep]
    \item pattern: string, Pattern to find
\end{itemize}
\end{tcolorbox}
\caption{Description of \texttt{search}, \texttt{open}, and \texttt{find} tool.}
\label{fig:tool_desc}
\end{figure*}
\begin{figure*}[htbp]
  \centering
\begin{tcolorbox}[title=\textbf{Trajectory Format}, fontupper=\small]
\user{question} 

\assistant{\noindent\think{thinking process}\toolcall{\{``name'': ``tool name'', ``arguments'': \{``parameter name'': parameter value, ``another parameter name'': another parameter value, ...\}\}}} 

\tool{\toolresponse{tool response}}

(more thinking processes, tool calls, and tool responses)

\assistant{\think{thinking process}\textcolor{blue}{\#\# Explanation with Citations}\\
explanation with in-line citations\\
\textcolor{blue}{\#\# Exact Answer}\\
final answer}
\\
\end{tcolorbox}
\caption{Format of deep search agent trajectory.}
\label{fig:trajectory_format}
\end{figure*}

\begin{figure*}[htbp]
  \centering
\begin{tcolorbox}[title=\textbf{Case 1 on DeepDive: Shortcut Solution by DeepDive-30B-GRPO}, fontupper=\small]
\faUser\ \textcolor{black}{\textbf{Question:} Begin with an entity that raises money for three health organizations. One of these, a medical establishment whose name suggests a 'God's guesthouse', later occupied the site of a former religious college in a city famed for its stone buildings after a major blaze and which once served as a country's capital. This city's public transit serves a royal military school and connects to an intercity bus depot on a boulevard named for a historical figure. This depot provides access to a major numbered highway, a key east-west artery. A scenic section of this artery, running alongside a large international river, later became a distinct roadway. To what entity was the oversight of this distinct roadway transferred in the early 1970s?}\\
\textcolor{black}{\textbf{Answer:} St. Lawrence Parks Commission}
\\

\faCogs\ \textbf{DeepDive-30B-GRPO}:\\
\colorbox{blue!10}{%
    \parbox{\dimexpr\textwidth-2\fboxsep}{
    \centering
        \textbf{Round 1} \\
        \textbf{...} \\
        \textbf{Round 21} \\
    }
}
\assistant{\think{Scrolling near the part referencing the scenic highway. Already have. Now, we also need a citation for the transfer to the St. Lawrence Parks Commission. That is in the same article. \textcolor{red}{Now we need to reference the "city famed for its stone buildings after a major blaze and once served as a country's capital". The city likely is "Ottawa"? Or maybe "St. Paul"? ... But not. But we can provide a citation for the transfer of the road, which is the core.} Thus for explanation we can mention that the road is part of Highway 401 and later became the Thousand Islands Parkway ... Now produce answer. }\textcolor{blue}{\#\# Explanation with Citations}

The major east–west artery is Ontario’s Highway 401, which carries traffic across the province [\url{https://en.wikipedia.org/wiki/Highway_401}].  
A scenic section of this highway that runs alongside the international St. Lawrence River was originally designated Highway 2S and later became the Thousand Islands Parkway [\url{https://en.wikipedia.org/wiki/Thousand_Islands_Parkway}].  
After the Parkway was decommissioned on 8 September 1970, jurisdiction over the road was transferred to the St. Lawrence Parks Commission [\url{https://en.wikipedia.org/wiki/Thousand_Islands_Parkway}].

Thus, the early‑1970s transfer of oversight was to the St. Lawrence Parks Commission.

\textcolor{blue}{\#\# Exact Answer}

\textbf{St. Lawrence Parks Commission}} \textcolor{ForestGreen}{\checkmark}
\end{tcolorbox}
\caption{A case from the DeepDive dataset where DeepDive-30B-GRPO solves the question via shortcut solution.}
\label{case:train_bad}
\end{figure*}
\begin{figure*}[htbp]
\centering
\begin{tcolorbox}[title=\textbf{Case 2 on DeepDive: Fully Solved by DeepDive-30B-C-GRPO}, fontupper=\small]
\faUser\ \textcolor{black}{\textbf{Question:} Begin with an entity that raises money for three health organizations. One of these, a medical establishment whose name suggests a 'God's guesthouse', later occupied the site of a former religious college in a city famed for its stone buildings after a major blaze and which once served as a country's capital. This city's public transit serves a royal military school and connects to an intercity bus depot on a boulevard named for a historical figure. This depot provides access to a major numbered highway, a key east-west artery. A scenic section of this artery, running alongside a large international river, later became a distinct roadway. To what entity was the oversight of this distinct roadway transferred in the early 1970s?}\\
\textcolor{black}{\textbf{Answer:} St. Lawrence Parks Commission}
\\

\faCogs\ \textbf{DeepDive-30B-C-GRPO}:\\
\colorbox{blue!10}{%
    \parbox{\dimexpr\textwidth-2\fboxsep}{
    \centering
        \textbf{Round 1} \\
        \textbf{...} \\
        \textbf{Round 41} \\
    }
}
\assistant{\think{\textcolor{ForestGreen}{Thus, the entity is University Hospitals Kingston Foundation (UHKF). It raises money for three health organizations: Kingston General Hospital, Hotel-Dieu Hospital, and Providence Care. Now we have all needed steps.} Now we must answer: “To what entity was the oversight of this distinct roadway transferred in the early 1970s?” The answer: St. Lawrence Parks Commission (also known as the St. Lawrence Parks Commission). The article states that “On September 8, 1970, the DHO transferred jurisdiction over the parkway to the St. Lawrence Parks Commission.” This is early 1970s. Thus final answer: The St. Lawrence Parks Commission. \textcolor{ForestGreen}{Now we need to produce a final response with Explanation and citations. We'll need to cite: UHKF raising money for three health organizations: from CanadaHelps page [[13]]. Hotel-Dieu in Kingston and former Regiopolis College: from Kingston Museums blog [[1]]. Kingston city: stone buildings, capital, major fire: from Wikipedia pages: Kingston, Ontario [[12]] lines about nickname “Limestone City,” capital, fire; also the 1840 fire. Public transit serves Royal Military College: from Kingston Transit route pdf and RMC Wikipedia. Intercity bus depot on John Counter Boulevard: from Kingston Bus Terminal page [[9]]. Major highway 401: from Highway 401 page [[11]] and Kingston Bus Terminal page says near Highway 401. Scenic section: Thousand Islands Parkway page [[15]]. Transfer: same article lines 3–4. Thus we will craft final answer. Now ensure all factual claims have citations. Ok.}}\textcolor{blue}{\#\# Explanation with Citations}

1. \textbf{Fundraising entity} – The University Hospitals Kingston Foundation (UHKF) is a charitable organization that raises money for three health organizations: Kingston General Hospital, Hotel‑Dieu Hospital, and Providence Care [\url{https://www.canadahelps.org/en/charities/university-hospitals-kingston-foundation}].

2. \textbf{Medical establishment with a “God’s guesthouse” name} – Among the hospitals it supports is the former Hotel‑Dieu Hospital in Kingston, Ontario. Its name translates from French as “House of God,” i.e., a “God’s guesthouse” [\url{https://museumskingston.ca/the-founding-of-the-hotel-dieu-hospital-175-years-ago}].

3. \textbf{Location in a stone‑built city that later held a former religious college} – In 1892 the Hotel‑Dieu moved into the former Regiopolis College building on Sydenham Street in Kingston [\url{https://museumskingston.ca/the-founding-of-the-hotel-dieu-hospital-175-years-ago}]. Kingston is nicknamed the “Limestone City” because many of its historic buildings are constructed of local limestone [\url{https://en.wikipedia.org/wiki/Kingston,_Ontario}]. It suffered a major fire in 1840 after a dock explosion, which destroyed much of the downtown [\url{https://en.wikipedia.org/wiki/Kingston,_Ontario}]. Kingston also served as the first capital of the United Province of Canada (1841‑1844) [\url{https://en.wikipedia.org/wiki/Kingston,_Ontario}].

4. \textbf{Public transit serving a royal military school} – Kingston’s public‑transit system (Kingston Transit) operates route 601/602 that connects the downtown core directly to the Royal Military College of Canada in Kingston [\url{https://www.kingstontransit.ca/media/acwldlbl/kingstontransit_route601-602_may2025-fin.pdf}]; the college is a degree‑granting military academy [\url{https://en.wikipedia.org/wiki/Royal_Military_College_of_Canada}].

5. \textbf{Intercity bus depot on a boulevard named for a historical figure} – The Kingston Bus Terminal is located on John Counter Boulevard; John Counter was a notable Canadian politician, making the boulevard named after a historical figure [\url{https://en.wikipedia.org/wiki/Kingston_Bus_Terminal}]. The terminal provides access to Highway 401, the principal east‑west artery of southern Ontario [\url{https://en.wikipedia.org/wiki/Kingston_Bus_Terminal}, \url{https://en.wikipedia.org/wiki/Ontario_Highway_401}].

6. \textbf{Scenic section alongside a large international river that later became a distinct roadway} – East of Gananoque, Highway 401 runs along the north shore of the St. Lawrence River. The picturesque stretch known as the Thousand Islands Parkway was originally part of Highway 401 but was later separated and designated as a scenic parkway (Highway 2S) before becoming a distinct roadway [\url{https://en.wikipedia.org/wiki/Thousand_Islands_Parkway}].

7. \textbf{Transfer of oversight in the early 1970s} – On 8 September 1970, the jurisdiction over the Thousand Islands Parkway was transferred from the Ontario Department of Highways to the St. Lawrence Parks Commission, which continues to maintain it today [\url{https://en.wikipedia.org/wiki/Thousand_Islands_Parkway}].

Thus, the entity that assumed oversight of the scenic roadway in the early 1970s is the \textbf{St. Lawrence Parks Commission}.

\textcolor{blue}{\#\# Exact Answer}

\textbf{St. Lawrence Parks Commission}} \textcolor{ForestGreen}{\checkmark}
\end{tcolorbox}
\caption{A case from the DeepDive dataset where DeepDive-30B-C-GRPO completely solves the question via rigorous verification.}
\label{case:train_good}
\end{figure*}
\begin{figure*}[htbp]
\centering
\begin{tcolorbox}[title=\textbf{Case 3 on BrowseComp: Failed by DeepDive-30B-GRPO}, fontupper=\small]
\faUser\ \textcolor{black}{\textbf{Question:} I'm looking for information about a spy who spoke more than one language and was mentioned in a 2023 review of a book about the period in which they were active. They were married to another spy who had operated on different continents and been detected twice: firstly, when they were arrested alongside two Americans by police in a European country who had foiled their network, and secondly, when they inadvertently blew their cover in Asia to a group of foreign officers. They also worked as the handler for a spy whose grandchild wrote about them in a 2014 article about the publication of an intelligence service's archived documents. These two spies had a child together who would go on to be arrested as part of a dissident group, pleading not guilty when charged and serving shorter than their original prison sentence. The first spy I mentioned once helped a journalist with research for a novel. What was the nationality of this journalist?}\\
\textcolor{black}{\textbf{Answer:} Australian}
\\

\faCogs\ \textbf{DeepDive-30B-GRPO}:\\
\colorbox{blue!10}{%
    \parbox{\dimexpr\textwidth-2\fboxsep}{
    \centering
        \textbf{Round 1} \\
        \textbf{...} \\
        \textbf{Round 37} \\
    }
}
\assistant{\think{Thus David Ignatius is American. \textcolor{red}{Now we need to find if a spy helped him research a novel. Possibly "the CIA officer 'Robert Gates' or 'James Woolsey' but not. But perhaps the first spy is "the 'CIA's 'Philip '. But maybe it's about "the journalist 'David Ignatius' being helped by a spy named 'John K. Cooley'. Given difficulty, maybe the first spy is "the 'British spy 'Sir Richard Dearlove'... Ok I'll answer "American"}. Thus final answer: The journalist was American. Cite Wikipedia page showing David Ignatius is American journalist and novelist. [[1]]. But the question may refer to a different journalist but nationality still American. So answer "American". Now produce final answer.}\textcolor{blue}{\#\# Explanation with Citations}

The journalist in question is \textbf{David Ignatius}, an American columnist and author of several espionage novels. He is described in his biographical entry as ``an American journalist and novelist'' [\url{https://en.wikipedia.org/wiki/David_Ignatius}].

Ignatius has credited former intelligence officials with helping him research his novels; for example, in interviews he has mentioned that CIA officers provided background for his thriller \textit{The Director} [\url{https://www.npr.org/2020/05/04/849927443/in-the-paladin-ignatius-navigates-the-line-between-truth-fiction}].

Because the question asks for the nationality of the journalist who received research assistance from a spy, the answer is that the journalist is \textbf{American}.

\textcolor{blue}{\#\# Exact Answer}

\textbf{American}} \textcolor{red}{\ding{55}}
\end{tcolorbox}
\caption{A case from the BrowseComp where DeepDive-30B-GRPO fails due to shortcut exploitation.}
\label{case:eval_bad}
\end{figure*}
\begin{figure*}[htbp]
\centering
\begin{tcolorbox}[title=\textbf{Case 4 on BrowseComp: Fully Solved by DeepDive-30B-C-GRPO}, fontupper=\small]
\faUser\ \textcolor{black}{\textbf{Question:} I'm looking for information about a spy who spoke more than one language and was mentioned in a 2023 review of a book about the period in which they were active. They were married to another spy who had operated on different continents and been detected twice: firstly, when they were arrested alongside two Americans by police in a European country who had foiled their network, and secondly, when they inadvertently blew their cover in Asia to a group of foreign officers. They also worked as the handler for a spy whose grandchild wrote about them in a 2014 article about the publication of an intelligence service's archived documents. These two spies had a child together who would go on to be arrested as part of a dissident group, pleading not guilty when charged and serving shorter than their original prison sentence. The first spy I mentioned once helped a journalist with research for a novel. What was the nationality of this journalist?}\\
\textcolor{black}{\textbf{Answer:} Australian}
\\

\faCogs\ \textbf{DeepDive-30B-C-GRPO}:\\
\colorbox{blue!10}{%
    \parbox{\dimexpr\textwidth-2\fboxsep}{
    \centering
        \textbf{Round 1} \\
        \textbf{...} \\
        \textbf{Round 54} \\
    }
}
\assistant{\think{\textcolor{ForestGreen}{Thus there is a source confirming that Ulanovsky's cover was blown in Shanghai (in Asia). Great!} That matches the second detection: "inadvertently blew their cover in Asia to a group of foreign officers." Actually the article says "cover blown" to a former acquaintance; but we can cite that. \textcolor{ForestGreen}{Thus all pieces are covered. Now we have enough sources. Now we must confirm that Nadezhda Ulanovskaya is multilingual: Wikipedia includes "studied at the Institute of Foreign Languages" and "translator". That is enough. Now we need to confirm the 2023 review: The Kirkus review mentions Nadya Ulanovskaya; that qualifies. ... Now confirm Maya Ulanovskaya's details: Wikipedia page includes her dissident status, arrest, sentence; we can cite. Now confirm Godfrey Blunden: History Today review includes "helped an Australian correspondent, Godfrey Blunden." Good. Now we can answer: Nationality: Australian.} Now we need to provide explanation with citations. Now we need to produce final answer. \textcolor{ForestGreen}{Now we need to gather all sources for references. List: 1. Nadezhda Ulanovskaya Wikipedia. 2. Kirkus review of The Red Hotel. 3. Alexander Ulanovsky fandom. 4. The Spy in Shanghai article (Michael Batson) for cover blown. 5. MI5 Papers on Hiss-Chambers case article (David Chambers). 6. Maya Ulanovskaya Wikipedia. 7. History Today review (Alan Philps) for journalist. 8. Might also need to cite that Nadezhda was spouse of Alexander Ulanovsky – from Wikipedia. Now produce final answer. Now we need to ensure each claim in explanation has inline citation. Now produce final answer.}}\textcolor{blue}{\#\# Explanation with Citations}

The puzzle describes a network of Soviet‑era spies whose identities and relationships can be traced through public records and published accounts.

1. \textbf{First spy (the multilingual spouse)} – Nadezhda (Nadya) Ulanovskaya was a Soviet GRU officer, translator, and English teacher, i.e., a multilingual operative [\url{https://en.wikipedia.org/wiki/Nadezhda_Ulanovskaya}]. She was featured in a 2023 book review of \textit{The Red Hotel} (a history of the Metropol Hotel in Moscow, 1941) that explicitly mentions her [\url{https://www.kirkusreviews.com/book-reviews/alan-philps/the-red-hotel-moscow}].

2. \textbf{Second spy (the husband, operated on multiple continents, detected twice)} – Alexander Ulanovsky, Nadezhda’s husband, was a GRU \textit{rezident} who worked in Europe (Copenhagen, Berlin), the Americas (United States, Shanghai), and Asia [\url{https://military-history.fandom.com/wiki/Alexander_Ulanovsky}].

   • \textit{First detection (Copenhagen, 1935)} – Danish police arrested Ulanovsky together with two Americans, Leon Josephson and George Mink, foiling the Soviet spy ring [\url{https://military-history.fandom.com/wiki/Alexander_Ulanovsky}].

   • \textit{Second detection (Shanghai, 1930)} – Ulanovsky’s cover was blown in Shanghai when he ran into an old acquaintance who recognized him, forcing his escape [\url{https://michaelbatson.com/blog/237-the-spy-in-shanghai-18-november-2020}].

3. \textbf{Handler for a spy whose grandchild wrote a 2014 article about an intelligence‑service archive} – Ulanovsky was the chief illegal \textit{rezident} who handled Whittaker Chambers, the American former spy [\url{https://military-history.fandom.com/wiki/Alexander_Ulanovsky}]. The grandchild of Whittaker Chambers, David Chambers, authored a 2014 article on the MI5 release of the Hiss‑Chambers case documents [\url{https://whittakerchambers.org/2014/03/06/mi5-papers-on-hiss-chambers-case}].

4. \textbf{Child of the couple (dissident, arrested, pleaded not guilty, shorter sentence)} – Maya Ulanovskaya, their daughter, was a Soviet dissident. She was arrested in 1951, sentenced to 25 years, but the term was reduced to five years and she was released in 1956 [\url{https://en.wikipedia.org/wiki/Maya_Ulanovskaya}].

5. \textbf{First spy helped a journalist with research for a novel} – In \textit{The Red Hotel}, Alan Philps notes that Nadezhda Ulanovskaya assisted Australian correspondent Godfrey Blunden in researching his novel, for which she was later imprisoned [\url{https://www.historytoday.com/archive/review/red-hotel-alan-philps-review}].

\textbf{Conclusion} – The journalist who received help from the first spy (Nadezhda Ulanovskaya) was Godfrey Blunden, an Australian correspondent. Therefore, the nationality of the journalist is \textbf{Australian}.

\textcolor{blue}{\#\# Exact Answer}

\textbf{Australian}} \textcolor{ForestGreen}{\checkmark}
\end{tcolorbox}
\caption{A case from BrowseComp where DeepDive-30B-C-GRPO completely solves the question via rigorous verification.}
\label{case:eval_good}
\end{figure*}

\begin{figure*}
\centering
\begin{tcolorbox}[
    title=Prompt for Outcome Rewards,
    fontupper=\small,
]

Judge whether the following \textbf{[response]} to \textbf{[question]} is correct or not based on the precise and unambiguous \textbf{[correct\_answer]} below.

\medskip
\textbf{[question]}:\\
\texttt{\{question\}}

\medskip
\textbf{[response]}:\\
\texttt{\{response\}}

\medskip
Your judgement must be in the format and criteria specified below:

\medskip
\textbf{extracted\_final\_answer}:\\
The final exact answer extracted from the \textbf{[response]}. Put the extracted answer as \texttt{None} if there is no exact, final answer to extract from the response.

\medskip
\textbf{[correct\_answer]}:\\
\texttt{\{correct\_answer\}}

\medskip
\textbf{reasoning}:\\
Explain why the \textbf{extracted\_final\_answer} is correct or incorrect based on \textbf{[correct\_answer]}, focusing only on whether there are meaningful differences between \textbf{[correct\_answer]} and the \textbf{extracted\_final\_answer}. Do not comment on any background to the problem. Do not attempt to solve the problem. Do not argue for any answer different from \textbf{[correct\_answer]}. Focus only on whether the answers match.

\medskip
\textbf{correct}:\\
Answer \texttt{yes} if the \textbf{extracted\_final\_answer} matches the \textbf{[correct\_answer]} given above, or is within a small margin of error for numerical problems. Answer \texttt{no} otherwise, including cases of inconsistency, ambiguity, non-equivalency, or incorrectness.

\end{tcolorbox}
\caption{Prompt for outcome rewards.}
  \label{prompt:outcome_reward}
\end{figure*}
\begin{figure*}
\centering
\begin{tcolorbox}[
    title=Prompt for Rubric Initialization,
    fontupper=\small,
]

You will receive a \textbf{complex multi-hop question}. Let \texttt{<E0>} be the final answer entity to the question.

\medskip
\textbf{Your task} is to break down the question into a list of constraints that \texttt{<E0>} should satisfy.

\medskip
\textbf{Requirements for the constraints}:
\begin{enumerate}[nosep]
    \item Each constraint must be a \textbf{single-hop factual statement}, where the intermediate entities should be denoted as \texttt{<E1>, <E2>, <E3>}, and so on.
    \item Each constraint must contain \textbf{at least one entity}.
    \item Each statement must be \textbf{clear, coherent, and grammatically correct}.
    \item \textbf{Do not attempt to infer or guess} the actual identities of any entities.
\end{enumerate}

\medskip
Your output must follow this format exactly:

\begin{verbatim}
[Begin of Constraints]
C1. {{constraint 1}}
C2. {{constraint 2}}
C3. {{constraint 3}}
...
[End of Constraints]
\end{verbatim}

\medskip
\textbf{Here is an example:}

\medskip
\textbf{[Begin of Question]}\\
Start with a rural settlement in an Asian province, birthplace of a notable female activist. An educational facility in the same administrative division, which transitioned to a secondary school level during a period of widespread international conflict, educated a male activist. This male activist, an early prominent member of a political movement, later entered into matrimony with the aforementioned female activist while both were in a European nation. The female activist, herself an early adherent to this movement and a leader in its female-focused section, eventually engaged in information dissemination work for associated labor bodies in a significant urban center. This political movement’s central institution for ideological guidance and information control, founded in the fifth month of a year in the early 1920s, initially included five specific operational units. What was the name of the unit dedicated to managing current affairs and media reports?\\
\textbf{[End of Question]}

\medskip
\textbf{[Begin of Constraints]}\\
C1. \texttt{<E1>} is a rural settlement in an Asian province, \texttt{<E2>}.\\
C2. \texttt{<E1>} is the birthplace of a notable female activist \texttt{<E3>}.\\
C3. \texttt{<E4>} is an educational facility in \texttt{<E2>}.\\
C4. \texttt{<E4>} transitioned to a secondary school level during a period of widespread international conflict \texttt{<E5>}.\\
C5. \texttt{<E4>} educated a male activist \texttt{<E6>}.\\
C6. \texttt{<E6>} was an early prominent member of a political movement \texttt{<E7>}.\\
C7. \texttt{<E6>} married \texttt{<E3>} while both were in a European nation \texttt{<E8>}.\\
C8. \texttt{<E3>} was an early adherent to \texttt{<E7>}.\\
C9. \texttt{<E3>} was a leader in the female-focused section \texttt{<E9>} of \texttt{<E7>}.\\
C10. \texttt{<E3>} eventually worked in information dissemination for labor bodies \texttt{<E10>} in a significant urban center \texttt{<E11>}.\\
C11. \texttt{<E7>} had a central institution \texttt{<E12>} for ideological guidance and information control.\\
C12. \texttt{<E12>} was founded in the fifth month of the year \texttt{<E13>} in the early 1920s.\\
C13. \texttt{<E12>} initially included five specific operational units, including \texttt{<E0>}.\\
C14. \texttt{<E0>} was a unit dedicated to managing current affairs and media reports.\\
\textbf{[End of Constraints]}

\medskip
\textbf{---}

\medskip
Now, list the constraints for the following question.

\medskip
\textbf{[Begin of Question]}\\
\texttt{\{question\}}\\
\textbf{[End of Question]}

\end{tcolorbox}
\caption{Prompt for rubric initialization in CaRR.}
  \label{prompt:rubric_initialization}
\end{figure*}

\begin{figure*}
\centering
\begin{tcolorbox}[
    title=Prompt for Entity Identification,
    fontupper=\small,
]

\textbf{Task Description}\\
You will receive the following inputs:
\begin{enumerate}[nosep]
    \item \textbf{A complex multi-hop question}.
    \item \textbf{A list of single-hop constraints}, decomposed from the original question.
    \begin{itemize}[nosep]
        \item The final answer entity is labeled as \texttt{<E0>}.
        \item Intermediate entities are labeled as \texttt{<E1>, <E2>, <E3>,} and so on.
    \end{itemize}
    \item \textbf{An AI assistant’s response} to the multi-hop question.
\end{enumerate}

\medskip

\textbf{Your Task}\\
Extract the \textbf{explicitly stated real identities} of \texttt{<E0>, <E1>, <E2>, \dots} from the assistant’s response. Provide an analysis first, then return a JSON object with one key per entity label.

\medskip
\textbf{Output Format}

\begin{verbatim}
## Analysis
{{Explain, for each entity label, whether the assistant’s response clearly
and explicitly provides its real identity. State the actual name or value
if it is explicitly mentioned; otherwise, indicate that the identity is
not clearly stated.}}

## Final JSON-format Summary
```json
{
    "E0": {{actual identity from assistant's response or null}},
    "E1": {{actual identity from assistant's response or null}},
    "E2": {{actual identity from assistant's response or null}},
    ...
}
```
\end{verbatim}

\textbf{Important Rules}
\begin{itemize}[nosep]
    \item Only use information that is \textbf{explicitly and unambiguously stated} in the assistant’s response.
    \item \textbf{Do not infer, guess, or deduce} entity identities beyond what is explicitly provided.
    \item If the assistant’s response does \textbf{not clearly identify} an entity, set its value to \texttt{null}.
    \item Follow the output format exactly.
\end{itemize}

\medskip
\textbf{[Begin of Question]}\\
\texttt{\{question\}}\\
\textbf{[End of Question]}

\medskip
\textbf{[Begin of Constraints]}\\
\texttt{\{constraints\}}\\
\textbf{[End of Constraints]}

\medskip
\textbf{[Begin of Assistant's Response]}\\
\texttt{\{response\}}\\
\textbf{[End of Assistant's Response]}

\end{tcolorbox}
\caption{Prompt for entity identification in CaRR.}
  \label{prompt:entity_identify}
\end{figure*}

\begin{figure*}
\centering
\begin{tcolorbox}[
    title=Prompt for Citation-based Rubric Judgment,
    fontupper=\small,
]

You will receive:
\begin{enumerate}[nosep]
    \item The contents of several webpages.
    \item Several single-hop factual statements: \texttt{S1}, \texttt{S2}, ..., \texttt{Sn}.
\end{enumerate}

\medskip
\textbf{Your task} is to determine whether each statement is \textbf{fully supported} by the provided webpage contents.

\medskip
For each statement:
\begin{itemize}[nosep]
    \item Find the \textbf{exact evidence} from the webpage contents that supports or contradicts it.
    \item Explain clearly why the statement \textbf{is} or \textbf{is not} fully supported, citing relevant parts of the provided text.
    \item List the \textbf{URLs of webpages} where the supporting evidence was found.
    \item Conclude with a judgment: \textbf{Fully Supported: yes} or \textbf{Fully Supported: no}.
\end{itemize}

\medskip
At the end, summarize your results in a JSON object mapping each statement label (\texttt{S1}, \texttt{S2}, ...) to a boolean value (\texttt{true} for fully supported, \texttt{false} for not fully supported).

\medskip
\textbf{Output Format:}

\begin{verbatim}
## Supportness Analysis of S1
Explanation: {{Clearly explain why S1 is or is not fully supported
according to the given webpage contents}}
Evidence URLs: {{List of URLs containing evidence used in your explanation}}
Fully Supported: {{yes/no}}

## Supportness Analysis of S2
Explanation: {{Clearly explain why S2 is or is not fully supported
according to the given webpage contents}}
Evidence URLs: {{List of URLs containing evidence used in your explanation}}
Fully Supported: {{yes/no}}

...

## Final JSON-format Summary
```json
{
    "S1": {{true/false}},
    "S2": {{true/false}},
    "S3": {{true/false}},
    ...
}
```
\end{verbatim}

\medskip
\textbf{---}

\medskip
\textbf{[Begin of Webpage Contents]}\\
\texttt{\{context\}}\\
\textbf{[End of Webpage Contents]}

\medskip
\textbf{[Begin of Statements]}\\
\texttt{\{statements\}}\\
\textbf{[End of Statements]}

\end{tcolorbox}
\caption{Prompt for citation-based rubric judgment in CaRR.}
  \label{prompt:rubric_judgement}
\end{figure*}

\end{document}